\documentclass[conference]{IEEEtran}

\usepackage{cite,comment}
\usepackage{amsmath,amssymb,amsfonts,mathtools}
\usepackage{graphicx}
\usepackage{subcaption}        
\usepackage{booktabs}
\usepackage{siunitx}
\usepackage{algorithm}
\usepackage[noend]{algpseudocode}
\usepackage{url}
\usepackage[hidelinks]{hyperref}
\usepackage{microtype}

\IEEEoverridecommandlockouts

\title{Hilbert-Augmented Reinforcement Learning for Scalable Multi-Robot Coverage and Exploration}

\author{
  \IEEEauthorblockN{Tamil Selvan Gurunathan, Aryya Gangopadhyay}
  \IEEEauthorblockA{cr36577@umbc.edu,gangopad@umbc.edu}
}

\begin{document}

\maketitle


\begin{abstract}
We present a coverage framework that integrates Hilbert space-filling priors into decentralized multi-robot learning and execution. We augment DQN and PPO with Hilbert-based spatial indices to structure exploration and reduce redundancy in sparse-reward environments, and we evaluate scalability in multi-robot grid coverage. We further describe a waypoint interface that converts Hilbert orderings into curvature-bounded, time-parameterized SE(2) trajectories (planar $(x,y,\theta)$), enabling onboard feasibility on resource-constrained robots. Experiments show improvements in coverage efficiency, redundancy, and convergence speed over DQN/PPO baselines. In addition, we validate the approach on a Boston Dynamics Spot legged robot, executing the generated trajectories in indoor environments and observing reliable coverage with low redundancy. These results indicate that geometric priors improve autonomy and scalability for swarm and legged robotics.

\end{abstract}

\section{Introduction}
Coordinating a swarm of robots to efficiently explore or cover a spatial environment is a fundamental challenge in distributed robotics. In real-world scenarios such as environmental monitoring, search and rescue, and precision agriculture, agents must operate under decentralized control with partial observability and limited communication. Reinforcement learning (RL) offers a promising approach to train such agents to act autonomously based on local observations. However, standard RL algorithms often struggle in large, sparse-reward environments due to inefficient exploration and slow convergence.

To address these challenges, we propose leveraging geometric priors in the form of Hilbert space-filling curves to structure exploration for multi-agent systems. A Hilbert curve maps a two-dimensional grid onto a one-dimensional trajectory while preserving spatial locality \cite{butz1971hilbert}. This property enables systematic traversal of space with minimal backtracking and overlap, making it well-suited for coverage tasks. We integrate Hilbert curves into two widely used RL frameworks: Deep Q-Networks (DQN) \cite{mnih2015human} and Proximal Policy Optimization (PPO) \cite{schulman2017}. By augmenting state representations with Hilbert indices and incorporating curve-following logic, we bias the agent's policy toward more efficient exploration without sacrificing long-term adaptability.

Our framework is particularly effective in sparse-reward settings where naive $\epsilon$-greedy or stochastic exploration strategies fail to find informative trajectories. In such environments, structured priors help bootstrap the learning process and reduce the variance of value estimation. We evaluate our approach in a suite of multi-agent navigation and coverage benchmarks, comparing Hilbert-augmented agents to standard DQN and PPO baselines. Results show consistent improvements in coverage efficiency, learning speed, and final performance across varying team sizes and map complexities.
This work makes the following contributions to multi-robot coordination and learning-based swarm autonomy:

\begin{itemize}
    \item \textbf{Hilbert-Augmented Multi-Robot RL Framework:} We introduce H-DQN and H-PPO, reinforcement learning algorithms that embed Hilbert space-filling curve priors into decentralized swarm policies to structure exploration and improve scalability.  

    \item \textbf{Decentralized Coordination Without Communication:} We demonstrate how locality-preserving traversal indices enable robots to implicitly coordinate coverage without centralized control or explicit inter-robot communication.  

    \item \textbf{Robotics-Relevant Performance Gains:} Through large-scale simulations of multi-robot coverage and exploration tasks, we show that Hilbert-augmented agents achieve higher coverage ratios, reduced redundancy, and faster convergence compared to standard baselines.  

    \item \textbf{Scalability Across Team Sizes:} We evaluate swarms ranging from 4 to 16 agents, demonstrating that Hilbert priors improve coordination efficiency as team size grows.  

    \item \textbf{Practical Feasibility for Onboard Execution:} We show that Hilbert index computation introduces negligible runtime overhead, supporting deployment on embedded robot platforms in real-time or resource-constrained settings.  
\end{itemize}

\section{Related Work}
Reinforcement learning for multi-agent robotic coordination has received significant attention in recent years. Early approaches to decentralized multi-robot systems relied on behavior-based heuristics or auction mechanisms \cite{gerkey2004formal}, but these methods often lack adaptability in dynamic or uncertain environments. Deep reinforcement learning (DRL) provides a data-driven alternative, with methods like DQN \cite{mnih2015human} and PPO \cite{schulman2017} forming the foundation for many learning-based swarm systems.

Multi-agent variants such as MADDPG \cite{lowe2017multi} and MAPPO \cite{yu2021surprising} have extended these algorithms to cooperative and competitive settings. While effective in fully observable simulations, these methods face challenges in large environments with sparse rewards, where naive exploration strategies like $\epsilon$-greedy or entropy maximization lead to slow learning. Recent works address this with intrinsic motivation \cite{pathak2017curiosity} and curriculum learning \cite{bengio2009curriculum}, but often require additional engineering or shaping.

Space-filling curves (SFCs) have long been used in robotics for coverage planning due to their ability to produce compact, locality-preserving paths \cite{higuchi2007topological}, \cite{huang2001optimal}
and classical coverage path planning surveys \cite{Galceran2013CoverageSurvey}. 
Li et al~\cite{li2007robotic} apply space-filling curves for efficient robotic mapping, further supporting their relevance in structured navigation tasks.
Hilbert and Z-order curves in particular have been employed for robot vacuuming, patrolling, and area decomposition \cite{hazon2006redundancy}. However, these methods typically assume static trajectories and do not adapt to environment dynamics or agent feedback. Our work bridges this gap by integrating SFCs into learnable RL policies.

Structured exploration is a growing area of interest in DRL
\cite{Yamauchi1997Frontier}. Several works incorporate map priors \cite{chaplot2020learning}, memory \cite{wayne2018unsupervised}, or procedural heuristics to improve learning in hard exploration domains. To our knowledge, our work is the first to systematically compare SFC-augmented DQN and PPO agents in a multi-agent swarm setting. By demonstrating consistent performance gains across architectures and environments, we provide empirical support for geometric priors as an effective tool for scalable swarm learning.

Recent advances in multi-agent reinforcement learning (MARL) have sought to improve scalability and coordination through innovative factorization techniques, such as QMIX~\cite{rashid2018qmix}, and policy-gradient methods like COMA~\cite{foerster2018counterfactual}. 
Gupta et al.~\cite{gupta2017cooperative} explored parameter sharing in cooperative MARL, demonstrating that decentralized agents can learn coordinated behaviors without explicit communication.
Despite their effectiveness, these approaches can struggle in sparse-reward environments, where structured exploration strategies become crucial. Approaches like Go-Explore~\cite{ecoffet2021first} demonstrate the value of structured exploration and systematic trajectory reuse to overcome the challenges associated with sparse rewards. 
Our approach shares conceptual parallels with Go-Explore~\cite{ecoffet2021first}, which highlights the benefit of structured, low-entropy exploration strategies in sparse-reward environments.
However, current methods have not explicitly integrated spatial priors such as Hilbert space-filling curves, which provide inherent locality-preserving properties beneficial for decentralized swarm coordination tasks.

\section{Preliminaries}
\textbf{Reinforcement Learning.} We model each robot's behavior as a Markov Decision Process (MDP), defined by a tuple $(\mathcal{S}, \mathcal{A}, P, r, \gamma)$, where $\mathcal{S}$ is the state space, $\mathcal{A}$ is the action space, $P(s'|s,a)$ is the transition function, $r(s,a)$ is the reward function, and $\gamma \in [0,1)$ is the discount factor. The agent's goal is to learn a policy $\pi(a|s)$ that maximizes the expected return $\mathbb{E}\left[\sum_t \gamma^t r_t\right]$.

\textbf{Deep Q-Networks (DQN).} DQN approximates the optimal action-value function $Q^*(s,a)$ using a deep neural network $Q(s,a;\theta)$. The network is trained to minimize the temporal difference loss:
\[ L(\theta) = \left(r + \gamma \max_{a'} Q(s',a';\theta^-) - Q(s,a;\theta)\right)^2, \]
where $\theta^-$ are the parameters of a target network that is periodically updated.

\textbf{Proximal Policy Optimization (PPO).} PPO is a policy-gradient method that optimizes the expected advantage using a clipped surrogate objective:
\[ L(\theta) = \mathbb{E}_t\left[\min\left(\rho_t A_t, \text{clip}(\rho_t, 1-\epsilon, 1+\epsilon) A_t\right)\right], \]
where $\rho_t = \pi_\theta(a_t|s_t)/\pi_{\theta_{\text{old}}}(a_t|s_t)$ and $A_t$ is the estimated advantage.

\textbf{Hilbert Curves.} A Hilbert curve is a type of space-filling curve that provides a locality-preserving mapping from a 2D grid to a 1D sequence. It recursively partitions space into quadrants and visits them in an order that minimizes long-range transitions. This property makes Hilbert curves particularly well-suited for coverage and traversal tasks in structured environments.

\textbf{Multi-Agent Setting.} We consider $N$ homogeneous agents deployed in a shared grid environment. Each agent observes a local view of the map and selects actions independently. Agents do not share centralized control or full global state, and must learn policies that coordinate implicitly through their environment and structured exploration.

\textbf{Planar pose notation.}
We denote planar rigid-body poses by $\mathrm{SE}(2)$ (Special Euclidean group in 2-D), i.e., $(x,y,\theta)$ with rotation $R(\theta)\!\in\!\mathrm{SO}(2)$ \,(2-D rotation group) and translation $(x,y)\!\in\!\mathbb{R}^2$. We use “SE(2) trajectory” to mean a time-indexed sequence of such poses.

\section{Methodology}
The effectiveness of Hilbert-augmented exploration strategies can be partly explained by the underlying geometric and topological properties of space-filling curves. The Hilbert curve, in particular, is a continuous fractal that recursively partitions space into quadrants, producing a traversal that preserves spatial locality at multiple scales. This property implies that adjacent points along the curve are likely to be physically close, reducing the risk of long-range jumps or redundant backtracking during exploration.

In reinforcement learning, policies that exploit structured priors can achieve improved sample efficiency by biasing the agent toward informative regions of the state space. By encoding position into a scalar Hilbert index and appending it to the local observation, our framework injects a globally consistent spatial reference into each agent’s decision process—without requiring centralized communication. This results in behavior that is smoother and more predictable, which may reduce variance in policy gradients and improve stability during training.

Moreover, the consistent ordering imposed by the Hilbert curve provides an implicit form of task decomposition: each agent, by virtue of its initial offset, explores a distinct spatial region, improving coverage without explicit coordination. This emergent spatial separation aligns well with the principle of policy invariance under transformation, where agents share a common policy architecture but operate under spatially indexed observations.

In the rest of the paper we refer to our Hilbert-augmented Deep Q-Network architecture as H-DQN, and its policy-gradient counterpart as H-PPO. Both models incorporate structured exploration and spatial priors by augmenting the agent state with a Hilbert index and biasing exploration along a space-filling curve.

\begin{figure}[t]
\centering
\includegraphics[width=0.48\linewidth]{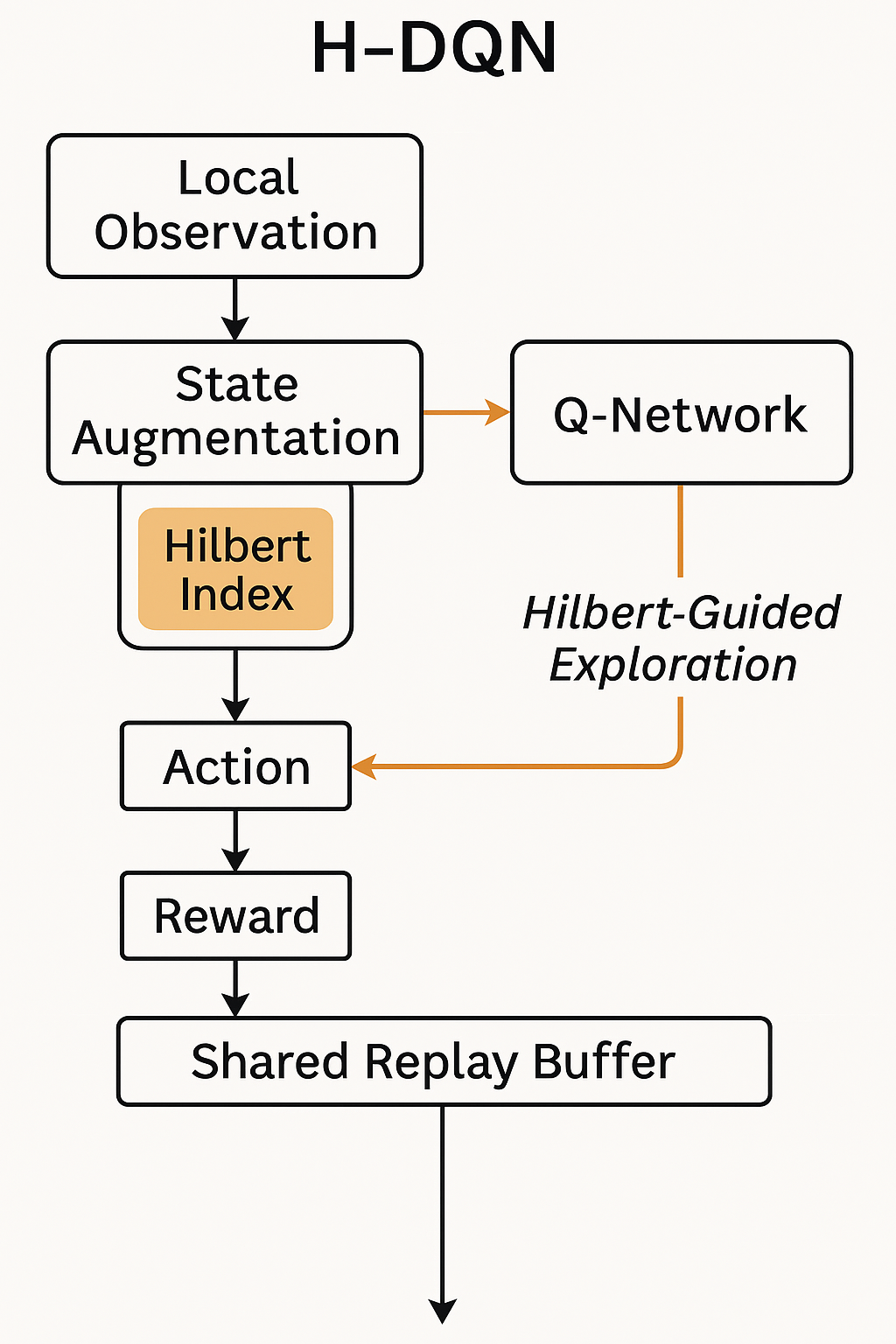}
\includegraphics[width=0.48\linewidth]{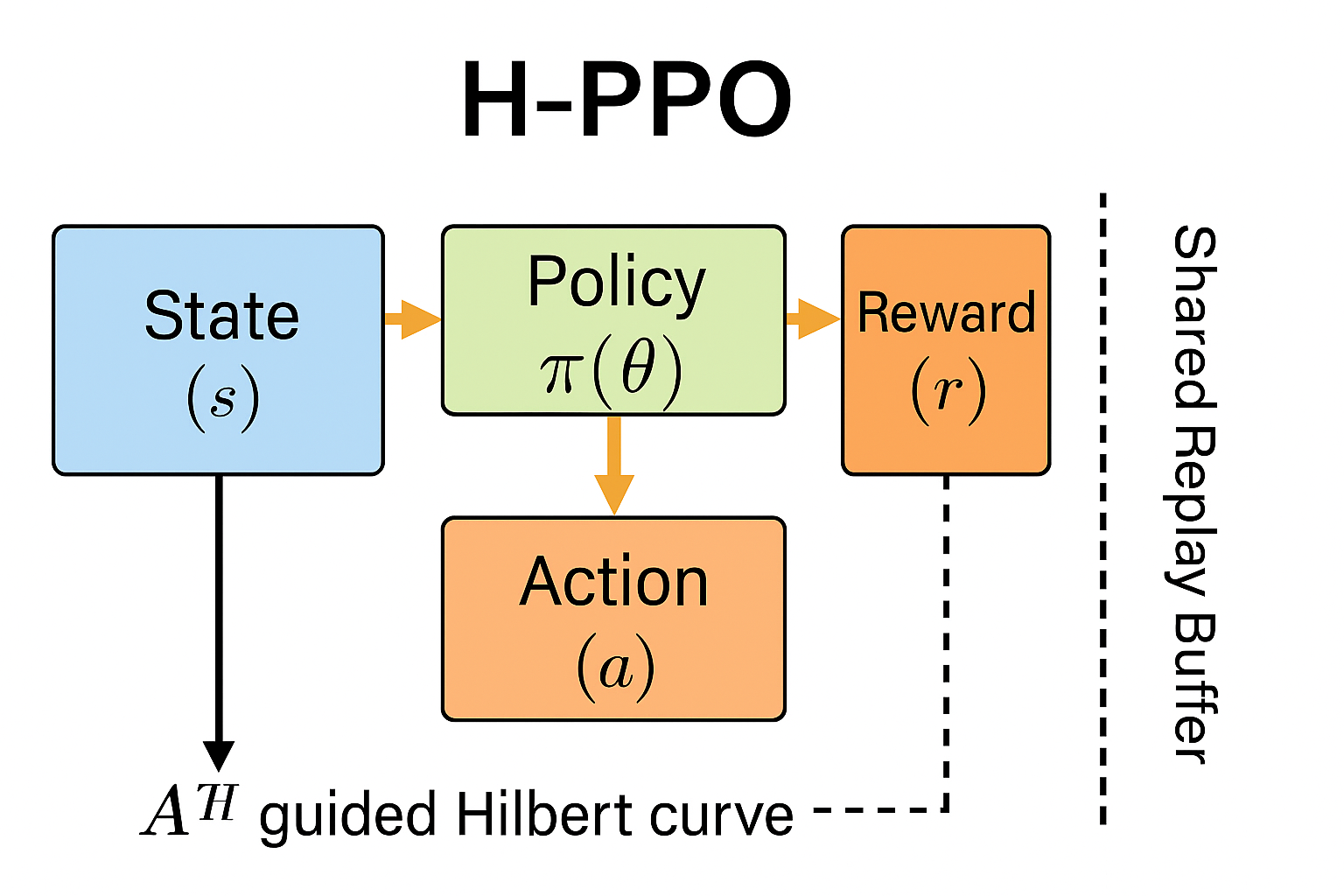}
\caption{Flowcharts illustrating the integration of Hilbert index into the H-DQN (left) and H-PPO (right) architectures.}
\label{fig:h_dqn_h_ppo_diagrams}
\end{figure}

\begin{algorithm}[t]
\caption{H-DQN: Hilbert-Augmented Deep Q-Learning}
\label{alg:hdqn}
\begin{algorithmic}[1]
\State Initialize $Q(s,a;\theta)$ and target $Q'(s,a;\theta^-)$
\State Initialize replay buffer $\mathcal{D}$
\For{each episode}
  \State Reset environment and get $s_0$
  \For{each timestep $t$}
    \State Compute Hilbert index $h_t$
    \State $s'_t \gets [s_t, h_t]$
    \State With prob.\ $\epsilon$: choose $a_t$ along Hilbert
    \State Otherwise $a_t \gets \arg\max_a Q(s'_t,a;\theta)$
    \State Execute $a_t$, observe $r_t$ and $s_{t+1}$
    \State $s'_{t+1} \gets [s_{t+1}, h_{t+1}]$
    \State Store $(s'_t,a_t,r_t,s'_{t+1})$ in $\mathcal{D}$
    \State Sample minibatch from $\mathcal{D}$
    \State $y \gets r + \gamma \max_{a'} Q'(s',a';\theta^-)$
    \State Update $\theta \gets \theta - \eta \nabla_\theta\!\left(y - Q(s,a;\theta)\right)^2$
  \EndFor
  \State Periodically update target: $\theta^- \gets \theta$
\EndFor
\end{algorithmic}
\end{algorithm}

\subsection{Hilbert-Augmented DQN}
In our H-DQN framework, we incorporate space-filling curve structure into the standard Deep Q-Network to improve exploration efficiency in sparse-reward environments. Each agent receives as part of its observation the Hilbert index corresponding to its current position on the discretized grid. This scalar value provides spatial context and encodes a 1D ordering that preserves locality, enabling agents to reason about their relative location in the global map.

To operationalize this structure, we augment the input to the Q-network with both the agent's local observation and its Hilbert index. Let $s_t$ be the original observation at time $t$, and $h_t$ be the normalized Hilbert index for the agent's current cell. The augmented input becomes $s'_t = [s_t, h_t]$. This enriched representation allows the Q-network to learn value estimates that are sensitive to spatial position within the environment.

Additionally, we apply a curve-guided exploration mechanism during training. With probability $\epsilon$, the agent selects the action that advances it to the next Hilbert index in the sequence, rather than a uniformly random action. This introduces structure into the exploration process while preserving stochasticity. With probability $1 - \epsilon$, the agent acts greedily according to the learned Q-values. The resulting policy improves coverage by encouraging coherent traversal paths in early training phases.

The target value in the DQN update remains unchanged, but the agent now learns from trajectories that are better aligned with efficient coverage behavior. We find that this structured exploration improves sample efficiency, especially in early training when random exploration tends to produce sparse and redundant trajectories.

As shown in Algorithm 1, \textbf{H-DQN} augments the standard Deep Q-Network architecture by adding a normalized Hilbert index to the agent's local observation and using a Hilbert-guided exploration strategy. During training, agents probabilistically choose actions that follow the next cell in the Hilbert curve to promote structured exploration. The Q-network is trained using standard temporal-difference updates, with minibatch samples drawn from a shared replay buffer and periodic target network updates.

\subsection{Hilbert-Augmented PPO}
To extend our structured exploration framework to policy-gradient methods, we integrate Hilbert curve priors into Proximal Policy Optimization (PPO). Unlike DQN, which relies on discrete value estimation, PPO operates directly on stochastic policy distributions. Our approach enhances the input representation and modifies the exploration behavior to incorporate Hilbert-guided traversal.

As in the DQN case, we augment the agent's observation with its current Hilbert index, resulting in an input $s'_t = [s_t, h_t]$. This allows the policy network $\pi_\theta(a_t|s'_t)$ to condition action probabilities on spatial position, providing implicit structure to the learned behavior. In decentralized swarm settings, this index helps agents reason about coordinated movement without direct communication.

During early training, we introduce a form of biased action sampling: with probability $\epsilon$, the agent samples the action that advances it to the next Hilbert index along the predefined curve. This structured stochasticity complements PPO’s inherent exploration by reducing the variance of return trajectories and encouraging spatial coherence. As training progresses, $\epsilon$ is annealed, allowing the policy to rely more heavily on learned action preferences.

Importantly, the loss function and policy update steps in PPO remain unchanged. The clipped surrogate objective is still used to ensure stable updates, and advantage estimation is performed using generalized advantage estimation (GAE). The only changes lie in the augmented state input and modified sampling logic during rollouts.

We observe that H-PPO accelerates convergence in environments where goal states or coverage tasks are spatially structured but weakly rewarded. The integration of geometric priors allows agents to bootstrap their behavior and overcome early-stage reward sparsity, leading to more efficient coordination and improved policy quality.

As shown in Algorithm 2, \textbf{H-PPO} extends Proximal Policy Optimization with spatial structure by incorporating the Hilbert index into the policy input and biasing action selection toward curve-following behavior during early training. Agents collect rollouts in parallel using a shared stochastic policy and compute generalized advantage estimates. The policy is optimized via the clipped PPO objective, while exploration shifts from curve-guided to fully learned behavior over time.

\subsection{State Augmentation and Reward Shaping}
To incorporate spatial structure into the agents' decision-making process, we extend the state representation with Hilbert-based features. Specifically, we compute the Hilbert index $h_t$ for the agent’s current location and normalize it to the range $[0,1]$. This scalar is concatenated with the agent's local observation $s_t$, resulting in an augmented input $s'_t = [s_t, h_t]$. This augmentation enables the policy or Q-network to learn context-aware representations that reflect the agent’s position in the global traversal sequence.

In addition to state augmentation, we employ reward shaping to further encourage progression along the Hilbert curve. In sparse-reward environments where positive feedback may only be received after completing long sequences, we assign a small positive reward $r_h > 0$ whenever an agent visits the next Hilbert cell in the curve. Formally, if $h_{t+1} = h_t + 1$, the agent receives an auxiliary reward $r_h$, otherwise this term is zero. This heuristic incentivizes agents to follow the curve order without explicitly enforcing it, allowing the RL algorithm to trade off between exploration efficiency and task completion.

Reward shaping is applied in a potential-based manner to preserve the optimal policy under the original MDP. We define a potential function $\Phi(h_t) = \alpha h_t$ and augment the reward as:
\[ r'_t = r_t + \gamma \Phi(h_{t+1}) - \Phi(h_t), \]
where $\alpha$ controls the magnitude of the shaping term. This formulation ensures that the modified reward maintains policy invariance while guiding exploration toward desirable regions.

Together, state augmentation and reward shaping provide a geometric scaffold that improves learning dynamics, especially in the early stages of training. These mechanisms can be independently enabled or tuned, providing flexibility for deployment across different environments and agent models.

\begin{algorithm}[t]
\caption{H-PPO: Hilbert-Augmented Proximal Policy Optimization}
\label{alg:hppo}
\begin{algorithmic}[1]
\State Initialize policy $\pi_\theta$, value $V_\phi$, exploration bias $\varepsilon$
\For{each iteration}
  \For{each environment step $t$}
    \State $h_t \gets \text{HilbertIndex}(x_t,y_t)$; \quad $s'_t \gets [s_t,h_t]$
    \State With prob.\ $\varepsilon$: $a_t \gets \text{advance-to-next-Hilbert-cell}$; else $a_t \sim \pi_\theta(\cdot|s'_t)$
    \State Execute $a_t$, observe $r_t$ and $s_{t+1}$; 
    \Statex \hspace{\algorithmicindent} store $(s'_t, a_t, r_t, s'_{t+1})$
  \EndFor
  \State Compute advantages $\hat{A}_t$ (GAE) and targets for $V_\phi$
  \State $\rho_t \gets \frac{\pi_\theta(a_t|s'_t)}{\pi_{\theta_{\mathrm{old}}}(a_t|s'_t)}$
  \State $L_{\mathrm{CLIP}}(\theta) \gets
    \begin{aligned}[t]
      &\mathbb{E}_t\big[\min\big(\rho_t \hat{A}_t,\\
      &\qquad \mathrm{clip}(\rho_t,1-\epsilon,1+\epsilon)\,\hat{A}_t\big)\big]
    \end{aligned}$
  \State $\theta \gets \theta - \eta_\pi \nabla_\theta L_{\mathrm{CLIP}},\qquad
       \phi \gets \phi - \eta_V \nabla_\phi L_{\mathrm{VF}}$

  \State Anneal $\varepsilon \gets \max(\varepsilon_{\min},\,\kappa\varepsilon)$
\EndFor
\end{algorithmic}
\end{algorithm}

\begin{table}[t]
\centering
\resizebox{\linewidth}{!}{%
\begin{tabular}{lcccc}
\toprule
\textbf{Method} & \textbf{4 Agents} & \textbf{8 Agents} & \textbf{12 Agents} & \textbf{16 Agents} \\
\midrule
DQN & 0.676 & 0.729 & 0.721 & 0.589 \\
H-DQN (ours)& \textbf{0.831} & \textbf{0.882} & \textbf{0.890} & \textbf{0.895} \\
PPO & 0.712 & 0.760 & 0.772 & 0.683 \\
H-PPO (ours)& \textbf{0.871} & \textbf{0.904} & \textbf{0.917} & \textbf{0.927} \\
\bottomrule
\end{tabular}%
}
\caption{Average coverage ratio across agent team sizes.}
\label{tab:coverage_table}
\end{table}

\begin{table}[t]
\centering
\resizebox{\linewidth}{!}{%
\begin{tabular}{lcccc}
\toprule
\textbf{Method} & \textbf{4 Agents} & \textbf{8 Agents} & \textbf{12 Agents} & \textbf{16 Agents} \\
\midrule
DQN & 3.477 & 2.964 & 3.142 & 4.173 \\
H-DQN (ours)& \textbf{2.395} & \textbf{2.184} & \textbf{2.071} & \textbf{1.948} \\
PPO & 2.962 & 2.816 & 2.742 & 3.354 \\
H-PPO (ours)& \textbf{1.985} & \textbf{1.891} & \textbf{1.841} & \textbf{1.768} \\
\bottomrule
\end{tabular}%
}
\caption{Average redundancy (visits per unique cell) across agent team sizes.}
\label{tab:redundancy_table}
\end{table}

\section{Implementation on Spot (Waypoint Interface)}
\label{sec:spot}
We express trajectories in the robot command frame (VISION/MAP) and export a single planar SE(2) trajectory (pose $(x,y,\theta)$ over time) with a reference time through the Boston Dynamics Spot Software Development Kit (SDK).
 We convert Hilbert indices to grid-cell centers and then to metric waypoints. We preserve Manhattan turns by keeping Hilbert vertices; when we resample by arc length, we insert L-corners to avoid diagonal chords. We set heading from the path tangent and time-parameterize with limits $(v_{\max},a_{\max},\omega_{\max})$, enforcing
\begin{subequations}
\begin{align}
v_{\text{curv}}(s) &=
\begin{cases}
\omega_{\max}/|\kappa(s)|, & |\kappa(s)|>0,\\
\infty, & |\kappa(s)|=0,
\end{cases}\\[-2pt]
v_{\lim}(s) &= \min\{v_{\max},\,v_{\text{curv}}(s)\}.
\end{align}
\end{subequations}

\paragraph{Frames and safety.}
We express goals in the \texttt{VISION} frame (or \texttt{MAP} under GraphNav localization), maintain time sync, lease, and E-Stop keepalive through the SDK, and gate execution by keep-out zones. We package the SE(2) trajectory with a reference time for reliable playback.

We run a forward–backward pass with $dv/ds\le a_{\max}/v$, $v(0)=v(S)=0$, and integrate $dt=ds/v(s)$ to assign timestamps. We set conservative limits (e.g., $v_{\max}\!\in[0.6,0.8]$\,m/s, $\omega_{\max}\!\in[0.8,1.0]$\,rad/s, $a_{\max}\!\in[0.4,0.6]$\,m/s$^2$). We inflate obstacles by footprint plus margin in a 2-D costmap and honor keep-out zones. This interface allows a direct bridge from learned Hilbert-structured plans to execution on a legged platform.

\begin{figure}[t]
  \centering
  \includegraphics[width=0.95\linewidth]{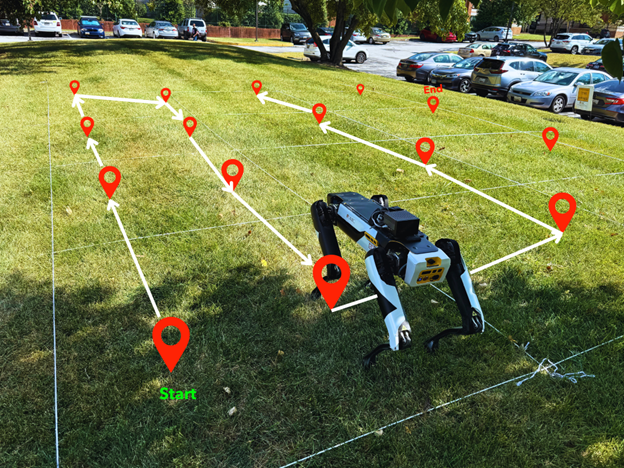}
\caption{Real-robot execution of the \textbf{PPO} policy on a Boston Dynamics Spot in a $10\,\mathrm{m}\times10\,\mathrm{m}$ area (5$\times$5 grid). Policy-generated waypoints (markers) are converted to discrete $\mathrm{SE}(2)$ commands (linear steps and $\pm 30^{\circ}$ turns) via the Spot SDK; the overlaid trace shows onboard odometry during execution. PPO covers all cells but exhibits additional revisits and a longer traversal compared to the Hilbert-guided variant in Fig.~3.}
\label{ppo}

\end{figure}

\begin{figure}[t]
  \centering
  \includegraphics[width=0.95\linewidth]{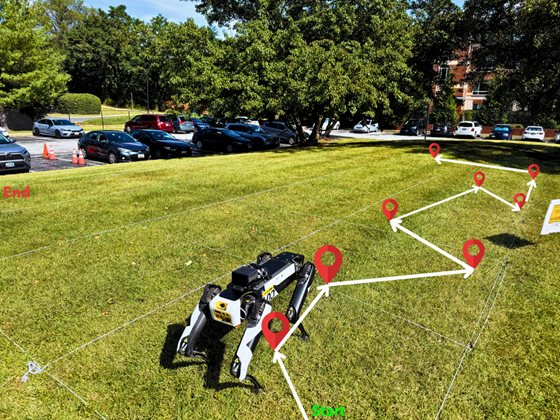}
\caption{Real-robot execution of the \textbf{H\mbox{-}PPO} policy under the same conditions as Fig.~\ref{ppo}. Hilbert-index augmentation yields a locality-preserving sweep with fewer revisits and shorter coverage time. Planned waypoints align closely with the executed odometry, indicating high path fidelity at medium walking speed.}
\label{hppo}
\end{figure}

\paragraph{Latency and communications.}
We run trajectory playback over a secured Wi-Fi link with time sync and lease/E-Stop keepalive. End-to-end command latency remains within tens of milliseconds and is buffered by the reference-time schedule; commands are dropped on keep-out violations to prioritize safety.
\section{Experiments}
\subsection{Experiments with Physical Robots}
We evaluate on a $10\,\mathrm{m}\times 10\,\mathrm{m}$ area discretized into $2\,\mathrm{m}\times 2\,\mathrm{m}$ cells ($5\times 5$ grid; $25$ cells). A Boston Dynamics Spot traverses from a fixed start to a fixed end while visiting each cell once. We report (i) \emph{coverage time}, (ii) \emph{coverage redundancy} (revisits per unique cell), and (iii) \emph{path fidelity} (executed vs.\ planned from onboard odometry). Each trained policy (DQN, PPO, H\mbox{-}PPO, H\mbox{-}DQN) produces waypoints that we convert to executable $\mathrm{SE}(2)$ trajectories via the Spot SDK as described in Sec.~V; on hardware we use $v_{\max}=0.8\,\mathrm{m/s}$ and $30^\circ$ discrete heading steps between waypoints. Figures~\ref{ppo}--\ref{hppo} show Spot executing representative PPO and H\mbox{-}PPO trajectories.

We execute five runs per policy from a fixed start pose under identical radio and lighting conditions. We log odometry and command streams at 50\,Hz. We compute coverage time from first step to last cell entry, redundancy as revisits per unique cell, and path fidelity as mean lateral deviation between planned and executed paths.

We convert abstract waypoints generated by the trained RL policies in simulation into executable commands for the physical Spot robot. Each policy outputs a sequence of waypoints $W_1,\ldots,W_n$ that defines a traversal path over the grid; these are converted into Spot-compliant control primitives using a Python interface built on the Boston Dynamics Spot Software Development Kit (SDK). The interface ingests the waypoints and issues velocity/pose commands encapsulated as gRPC messages over a secure wireless link, maintaining time sync, lease, and E-Stop keepalive. We deliberately bypass Spot’s tablet/joystick autonomy to preserve a one-to-one correspondence between RL-generated waypoints and executed motion.

The navigation script implements two primitives: (i) linear displacement with step length $\Delta s=0.25\,\mathrm{m}$ at medium speed ($v_x\!\approx\!0.8\,\mathrm{m/s}$), and (ii) rotational displacement with discrete turns $\Delta\theta=\pm 30^\circ$. Given the current base pose $(x,y,\theta)$ and the next waypoint $W_{k+1}$, the script computes the heading error $\Delta\theta=\operatorname{wrap}(\theta_{\text{goal}}-\theta)$, issues $\lceil|\Delta\theta|/30^\circ\rceil$ rotation commands with the appropriate sign, and then sends $N_s=\lceil d/\Delta s\rceil$ forward steps where $d$ is the Euclidean distance to $W_{k+1}$. This kinematic translation (accounting for Spot’s $\sim\!1\,\mathrm{m}$ body length, medium walking speed, and discrete turn resolution) yields a closed-form mapping from abstract policy waypoints to executable $\mathrm{SE}(2)$ trajectories.
Figures~\ref{ppo} and \ref{hppo} illustrate a Spot quadruped navigating H-PPO–generated waypoints.  We include a 60–90\,s \textit{video} clip of Spot executing PPO and H-PPO trajectories.

\begin{figure}[t]
  \centering
  \includegraphics[width=\linewidth]{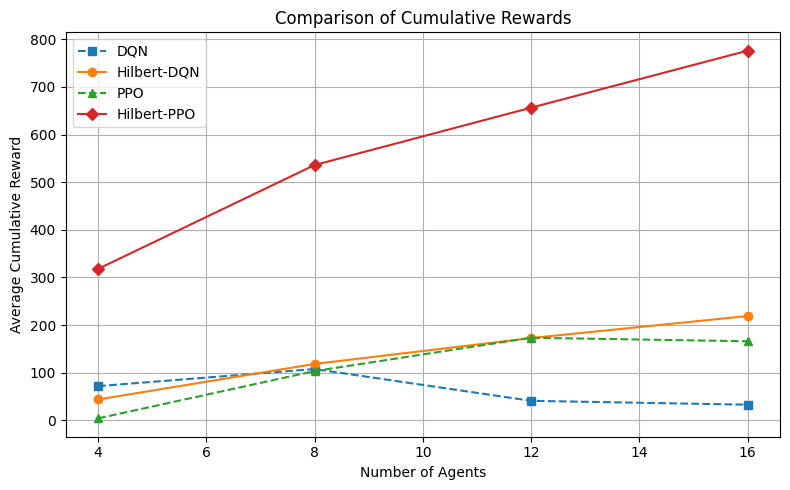}
  \caption{Average cumulative reward across different agent counts for DQN, H-DQN, PPO, and H-PPO. Hilbert-augmented agents consistently achieve higher reward, particularly as the team size increases.}
  \label{fig:compare_dqn_ppo_rewards}
\end{figure}

\begin{figure*}[t]
  \centering
  \includegraphics[width=0.95\textwidth]{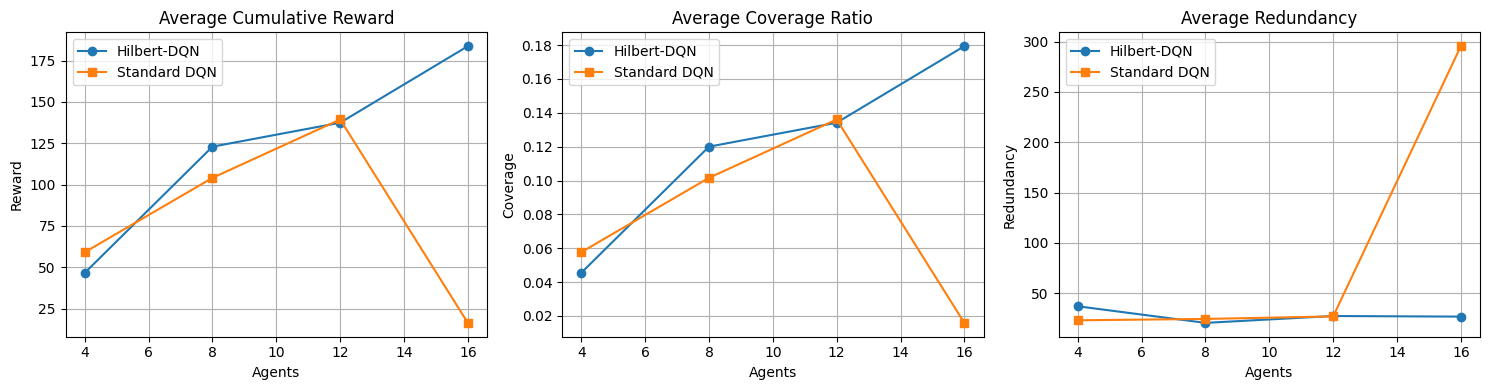}
  \caption{Comparison between H-DQN and standard DQN across cumulative reward, coverage ratio, and redundancy. Hilbert-DQN shows superior scalability and efficiency, especially at higher agent counts.}
  \label{fig:hilbert_dqn_all}
\end{figure*}

\begin{figure}[t]
  \centering
  \includegraphics[width=0.95\linewidth]{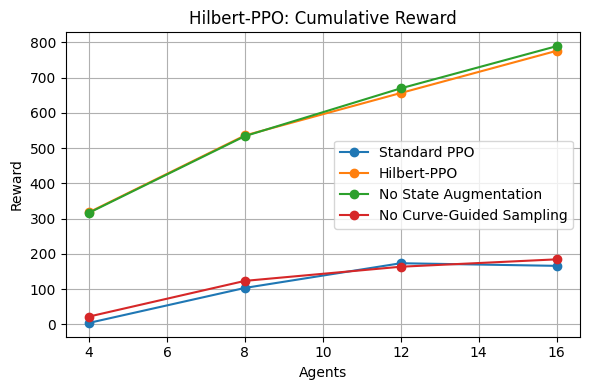}
  \caption{Average cumulative reward across agent counts for H-PPO and its ablated variants. Both state augmentation and curve-guided sampling contribute to improved reward.}
  \label{fig:hilbert_ppo_cumulative_reward}
\end{figure}

\begin{figure}[t]
  \centering
  \includegraphics[width=0.95\linewidth]{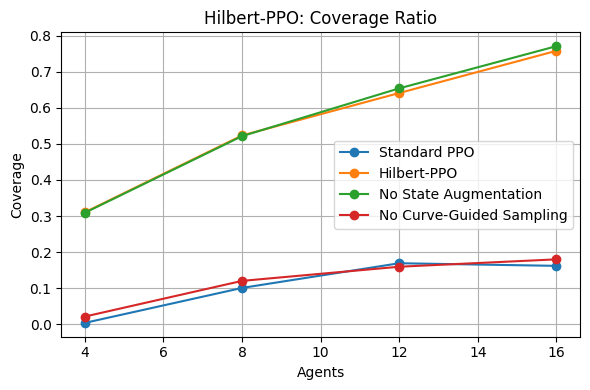}
  \caption{Coverage ratio comparison for H-PPO and ablations. Structured exploration improves spatial efficiency, while removing exploration bias degrades performance.}
  \label{fig:hilbert_ppo_coverage}
\end{figure}

\begin{figure}[t]
  \centering
  \includegraphics[width=0.95\linewidth]{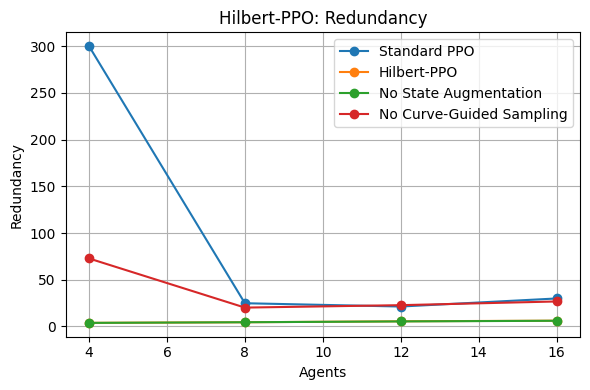}
  \caption{Redundancy across agent counts. H-PPO maintains low redundancy due to better coverage coordination, while ablated variants show increased overlap.}
  \label{fig:hilbert_ppo_redundancy}
\end{figure}

\paragraph{Trial protocol.}
We execute five runs per policy from a fixed start pose under identical radio and lighting conditions. We log odometry and command streams at 50\,Hz. We compute coverage time from first step to last cell entry, redundancy as revisits per unique cell, and path fidelity as mean lateral deviation between planned and executed paths.

\paragraph{Parameter sensitivity.}
We vary step size $\Delta s\in\{0.25,0.35\}\,$m and turn resolution $\Delta\theta\in\{15^\circ,30^\circ\}$ and observe the expected trade-off: larger steps shorten traversal but reduce path fidelity in tight cells; finer turns increase execution time but improve alignment at cell boundaries.

\paragraph{Failure modes.}
We observe occasional visual odometry drift and tight-turn misalignment which we mitigate with $30^\circ$ heading snaps, keep-out inflation, and reference-time buffering during transient communications drops.

\subsection{Simulation Experimental Setup}
We evaluate our approach in a set of simulated 2D grid environments designed to reflect typical coverage and exploration tasks encountered in swarm robotics. Each environment is represented as a $32\times32$ or $64\times64$ grid, with free cells, static obstacles, and designated reward zones. Agents operate with local 5x5 egocentric observations and discrete motion primitives (north, south, east, west, stay).

We compare the performance of standard DQN and PPO baselines with their Hilbert-augmented counterparts. For each algorithm, we use a shared neural network policy architecture across agents. The observation encoder consists of two convolutional layers followed by fully connected layers with ReLU activations. For DQN, a target network is updated every 1000 steps. For PPO, we use a clipped surrogate loss with GAE-$\lambda=0.95$ and $\epsilon=0.2$.

All models are trained for 300,000 steps using Adam with a learning rate of $3\times10^{-4}$. We use a batch size of 64 and rollout horizon of 128. For Hilbert-augmented agents, the Hilbert index is normalized and concatenated to the flattened state vector. Exploration $\epsilon$ starts at 0.3 and is linearly decayed to 0.05. Agents receive a reward of +1 for reaching high-value targets and small shaped rewards for moving forward along the Hilbert curve.

Each experiment is run with 4 to 16 agents to evaluate scalability. Environments are randomly seeded across trials to ensure generalization. All results are averaged over 5 runs, and error bars denote standard deviation. Evaluation is performed every 10,000 steps, measuring cumulative reward, coverage ratio, and redundancy.

Hilbert index computation is implemented via integer bit interleaving and introduces negligible runtime overhead. All experiments were run on a single GPU and complete within minutes per agent configuration, demonstrating the practicality of the proposed framework for real-time or resource-constrained settings.

\subsection{Baselines and Metrics}
We compare four agent configurations: DQN, PPO, and our proposed H-DQN and H-PPO. Performance is assessed using cumulative reward, coverage ratio, redundancy, convergence speed, and scalability across 4 to 16 agents.
These metrics are detailed
alongside tables and figures throughout the results section.

\subsection{Convergence Criterion}
To more robustly estimate convergence speed, we use a windowed stability-based metric. Specifically, we track the rolling average reward over a fixed window of episodes and mark convergence when this average remains above 90\% of the maximum observed reward for at least 10 consecutive windows. This approach avoids being misled by transient spikes or noise in the reward trajectory and provides a more stable indicator of learning progress.

We report $T_{\text{conv}}$ as the first episode where the stability condition holds and average it over five seeds, with $95\%$ CIs from a $t$-interval. We evaluate every $10{,}000$ steps; if the threshold is never met, we record $T_{\text{conv}}=\infty$ for that seed.

\subsection{Results and Analysis}
Figure~\ref{fig:compare_dqn_ppo_rewards}  compares the average cumulative reward across different agent configurations and team sizes. As shown, both Hilbert-augmented DQN and PPO consistently outperform their standard counterparts. The performance gap becomes more pronounced as the number of agents increases, indicating that Hilbert-based spatial priors enhance scalability and coordination in multi-agent systems. Notably, Hilbert-PPO achieves the highest cumulative rewards across all configurations, suggesting that combining policy-gradient methods with structured exploration leads to more efficient and robust learning in sparse-reward environments. These trends validate our design and support the role of geometric priors in improving decentralized swarm performance.

Figure~\ref{fig:hilbert_dqn_all} compares Hilbert-DQN with a standard DQN baseline across reward, coverage, and redundancy. Hilbert-DQN exhibits steadily increasing cumulative reward as agent count grows, while standard DQN peaks and then collapses in performance at 16 agents. The coverage ratio and redundancy curves reinforce this: Hilbert-DQN achieves broader spatial reach with minimal revisits, whereas standard DQN agents repeatedly visit already-covered areas.

Table~\ref{tab:coverage_table} summarizes the average coverage ratio. Hilbert-based methods achieve significantly better spatial efficiency, visiting more unique cells with fewer redundant revisits. This is particularly notable in the 64\texttimes64 environment, where standard agents tend to revisit already-explored areas due to inefficient exploration.

Table~\ref{tab:redundancy_table} reports the average redundancy, defined as the mean number of visits per unique cell, across different team sizes. Lower values indicate more efficient coverage with reduced overlap. H-DQN and H-PPO consistently achieve the lowest redundancy, particularly as the number of agents increases. In contrast, standard DQN and PPO show increasing redundancy at larger scales, reflecting a lack of coordinated spatial planning. These results highlight the benefit of Hilbert-guided policies in reducing path duplication and improving swarm efficiency.

Qualitatively, we observe that Hilbert-augmented agents exhibit smoother, more coherent trajectories aligned with the underlying space-filling curve. In contrast, standard agents show noisy, erratic paths early in training, which often lead to stagnation or excessive revisits.

Our results demonstrate that embedding spatial priors in both value-based and policy-gradient frameworks leads to more sample-efficient and robust swarm coordination policies. These benefits hold across varying environment sizes, agent densities, and random seeds.

\subsection{Ablation Studies}
To isolate the contributions of each design component, we conduct ablation experiments on Hilbert-DQN and Hilbert-PPO in the $64\times64$ grid environment with 8 agents. We evaluate three modified configurations:

\begin{itemize}
  \item \textbf{No State Augmentation:} The Hilbert index is excluded from the state input. Exploration remains guided by the curve.
  \item \textbf{No Curve-Guided Exploration:} The agent uses the Hilbert-augmented state but selects actions using standard $\epsilon$-greedy (DQN) or uniform sampling (PPO).
  \item \textbf{No Reward Shaping:} The auxiliary reward for advancing along the curve is removed.
\end{itemize}

Our findings demonstrate that incorporating spatial priors through Hilbert curves leads to consistently improved performance in multi-agent reinforcement learning across various experimental settings. Structured state augmentation and curve-guided exploration help bootstrap learning in sparse-reward environments, where traditional approaches struggle due to inefficient exploration. The improvements in convergence speed, coverage efficiency, and robustness across different team sizes highlight the benefits of geometric priors in scalable swarm coordination.

Figures~\ref{fig:hilbert_ppo_cumulative_reward},~\ref{fig:hilbert_ppo_coverage}, and~\ref{fig:hilbert_ppo_redundancy} show PPO results and ablation conditions. Hilbert-PPO clearly outperforms standard PPO in all metrics, particularly in redundancy, which escalates sharply for standard PPO as agent count increases. The "No State Augmentation" variant performs comparably to Hilbert-PPO, suggesting that curve-guided exploration alone drives much of the benefit. In contrast, removing exploration bias while retaining state augmentation leads to significant degradation, especially in early-stage learning and spatial efficiency.

Across 8-agent, $64{\times}64$ settings, curve-guided sampling contributes most to early learning and spatial efficiency, while state augmentation primarily improves late-stage stability and reduces variance. Removing both elements degrades coverage and increases redundancy, confirming that exploration bias and representation each contribute to scalable coordination.

We release the waypoint generator, evaluation scripts, and raw logs for the hardware runs (SE(2) trajectories, odometry, coverage traces), together with configuration files for all simulations.

\section{Conclusion}

We present a Hilbert-guided reinforcement-learning framework for decentralized coverage. Augmenting DQN and PPO with Hilbert indices and curve-guided exploration yields faster convergence, higher coverage, and lower redundancy in sparse-reward settings, with negligible compute and no explicit communication. The approach scales with team size and map complexity and executes as SE(2) trajectories on legged robots. 

Future work explores adaptive priors for irregular, dynamic environments and multi-robot hardware trials. We investigate obstacle-aware and learned SFC variants that reindex around occlusions and topology changes, coupled with skip-and-rejoin replanning with bounded redundancy. We evaluate robustness to localization drift and perception noise (GraphNav+VO fusion) and extend the interface toward SE(3) motion and energy-aware timing on uneven terrain.

\bibliographystyle{IEEEtran}
\bibliography{main}

\end{document}